\documentclass[letterpaper, 10 pt, journal, twoside]{IEEEtran}  %

\usepackage{lipsum}
\usepackage{subcaption}
\usepackage{graphics} %
\usepackage{graphicx}
\usepackage{epsfig} %
\usepackage{amsmath} %
\usepackage{amssymb}  %
\usepackage{amsfonts}

\usepackage{bm}

\usepackage{siunitx}
\usepackage{xcolor}
\usepackage[breaklinks]{hyperref}
\usepackage[capitalise]{cleveref}

\usepackage{booktabs}
\usepackage{threeparttable}
\usepackage{makecell}
\usepackage{multirow}

\usepackage[nolist,nohyperlinks]{acronym}
\acrodef{method}[AOM]{ACRONYM OF METHOD}
\acrodef{gnss}[GNSS]{Global Navigation Satellite System}
\acrodef{icp}[ICP]{Iterative Closest Point}
\acrodef{uas}[UAS]{Unmanned Aerial Systems}
\acrodef{ransac}[RANSAC]{Random Sample Consensus}
\acrodef{slam}[SLAM]{Simultaneous Localization And Mapping}
\acrodef{fmcw}[FMCW]{Frequency Modulated Continuous Wave}
\acrodef{pca}[PCA]{Principal Component Analysis}
\acrodef{ekf}[EKF]{Extended Kalman Filter}
\acrodef{isam2}[iSAM2]{Incremental Smoothing and Mapping}
\acrodef{rio}[RIO]{Radar Inertial Odometry}
\acrodef{rmse}[RMSE]{Root Mean Square Error} 
\acrodef{ape}[APE]{Absolute Pose Error}
\acrodef{cfar}[CFAR]{Constant False Alarm Rate}
\acrodef{snr}[SNR]{Signal to Noise Ratio}
\acrodef{rcs}[RCS]{Radar Cross Section}
\acrodef{imu}[IMU]{Inertial Measurement Unit}
\acrodef{rdmap}[RD-Map]{Range-Doppler Map}
\acrodef{cacfar}[CA-CFAR]{Cell-Averaging \acs{cfar}}
\acrodef{cut}[CUT]{Cell Under Test}
\acrodef{jpl}[JPL]{Jet Propulsion Laboratory}
\acrodef{fft}[FFT]{Fast-Fourier Transform}
\acrodef{fov}[FoV]{Field of View}
\acrodef{if}[IF]{Intermediate Frequency}

\DeclareMathOperator*{\argmin}{argmin}

\usepackage{nth}

\def\hlfon{0} %
\def\hlf#1{%
  \ifnum\hlfon=1
    \textcolor{blue}{#1}%
  \else
    #1%
  \fi
}

\usepackage[symbol]{footmisc}

\title{
Robust High-Speed State Estimation for Off-road Navigation using Radar Velocity Factors}

\author{Morten~Nissov$^{1,2}$, Jeffrey~A.~Edlund$^1$, Patrick~Spieler$^1$, Curtis~Padgett$^1$, Kostas~Alexis$^2$, and Shehryar~Khattak$^1$%
\thanks{Manuscript received: June, 9, 2024; Revised August, 14, 2024; Accepted October, 6, 2024.}%
\thanks{This paper was recommended for publication by Editor Pascal Vasseur upon evaluation of the Associate Editor and Reviewers' comments.}%
\thanks{This work was carried out at the Jet Propulsion Laboratory, California Institute of Technology, under a contract with the National Aeronautics and Space Administration (80NM0018D0004). This work was partially supported by Defense Advanced Research Projects Agency (DARPA). This work was also partially funded by the Research Council of Norway Award NO-321435. (Corresponding author: Morten Nissov {\tt\footnotesize morten.nissov@ntnu.no})}%
\thanks{\copyright 2024. California Institute of Technology. Government sponsorship acknowledged. All rights reserved.}%
\thanks{$^1$ NASA Jet Propulsion Laboratory, California Institute of Technology, Pasadena, CA, USA}%
\thanks{$^2$ Autonomous Robots Lab at the Norwegian University of Science and Technology, Trondheim, Norway}%
\thanks{Digital Object Identifier (DOI): see top of this page.}
}

\begin{document}

\maketitle

\begin{abstract}
Enabling robot autonomy in complex environments for mission critical application requires robust state estimation. Particularly under conditions where the exteroceptive sensors, which the navigation depends on, can be degraded by environmental challenges thus, leading to mission failure. 
It is precisely in such challenges where the potential for \acf{fmcw} radar sensors is highlighted: as a complementary exteroceptive sensing modality with direct velocity measuring capabilities.
In this work we integrate radial speed measurements from a FMCW radar sensor, using a
radial speed factor, to provide linear velocity updates into a sliding--window state estimator for fusion with LiDAR pose and \acs{imu} measurements. 
We demonstrate that this augmentation increases the robustness of the state estimator to challenging conditions present in the environment and the negative effects they can pose to vulnerable exteroceptive modalities. The proposed method is extensively evaluated using robotic field experiments\footnote[2]{\url{https://youtu.be/reZEgeataBk?si=SS2XNXlMOPEbgsI4}} conducted using an autonomous, full-scale, off-road vehicle operating at high-speeds (\SI{\sim12}{\meter\per\second}) in complex desert environments. Furthermore, the robustness of the approach is demonstrated for cases of both simulated and real-world degradation of the LiDAR odometry performance along with comparison against state-of-the-art methods for radar-inertial odometry on public datasets.
\end{abstract}
\begin{IEEEkeywords}
\hlf{Field Robots; Sensor Fusion; Localization}
\end{IEEEkeywords}

\section{Introduction}
\IEEEPARstart{M}{odern} approaches for autonomous robotic operation, especially under challenging conditions, require trustworthy state estimation to facilitate both low- and high-level tasks such as control, planning, and decision making. Such is clear from~\cite{cadena2016future,ebadi2024present}, where approaches tend to follow a typical architecture: inertial navigation aided by exteroceptive sensing acting as a replacement for GPS in the standard aided inertial navigation system. This class of approaches arises from the necessity for autonomy in situations with limited to no GPS access (e.g. underground, poor weather, etc). 

\begin{figure}[t!]
    \centering
    \includegraphics[width=\linewidth]{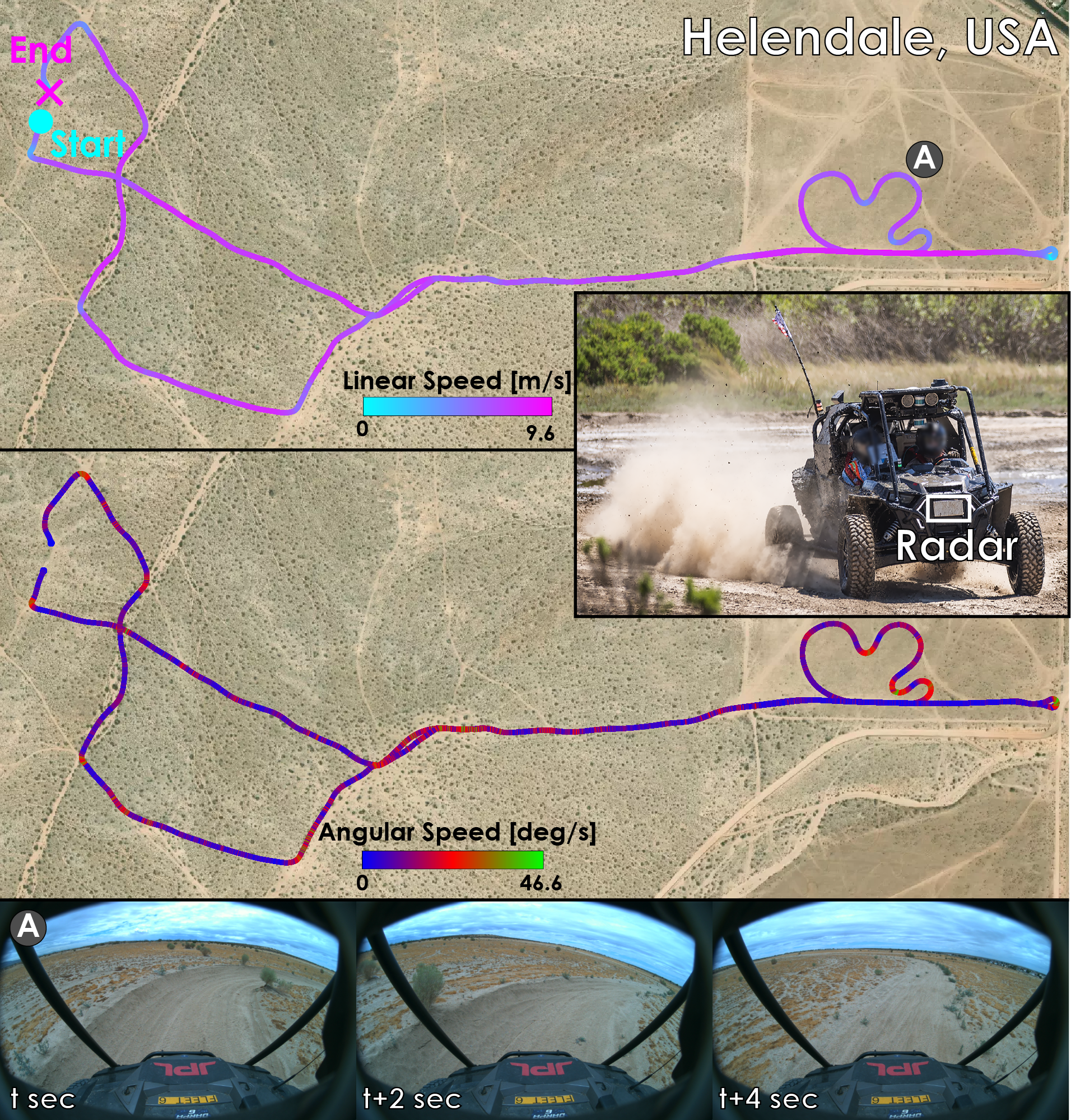}
    \caption{Figure shows the robotic vehicle during a field test conducted in a desert environment. The \SI{4}{\kilo\meter} trajectory is annotated with linear and rotational speeds achieved during the experiment. The barren and structure--less environment proved challenging for LiDAR odometry and lead to failure during sharp turns (marked by A). Bottom row shows the environment through images captured by on-board cameras.}
    \label{fig:intro:figure}
\vspace{-2em}
\end{figure}

Typically, pre-integrated IMU measurements and exteroceptive pose updates are sufficient.
Although, in such architectures the consistent, high-rate velocity estimates needed for robotic control necessarily come from integration of the accelerometer (with angular velocity provided by the gyroscope).
For applications at high-speeds, in complex environments, and under challenging conditions, maintaining accurate but high-rate linear velocity estimates in addition to pose becomes increasingly important for robust operation.
For pose estimation visual- and LiDAR-odometry can provide reliable estimates, but for velocity estimation inclusion of linear velocity inputs aiding the accelerometer integration can be beneficial. Especially if velocity updates are provided at a higher rate than the visual- or LiDAR-odometry updates. For this purpose, radar sensors are of interest, given that they can sample at high rate (up to \SI{60}{\hertz}) and provide direct linear velocity measurements. Furthermore, combining the radar with typical exteroceptive pose updates means that the state estimator receives measurements on both pose and linear velocity, improving the overall robustness of the estimator and leaving only IMU biases to be estimated without a direct measurement. This configuration resembles more closely the classical GPS-aided inertial navigation solutions for ground vehicles, in which velocity measurements were typically provided by wheel-odometry for dead-reckoning~\cite{gpsIMUwheel}.
However, to take full advantage of these high-rate direct linear velocity measurements, they need to be incorporated into the state estimator in a low-latency manner.

Motivated by the discussion above, this work presents a method aiming to improve state estimation robustness by integration of radial speed measurements from a \ac{fmcw} radar. 
The proposed method emphasizes low-latency updates and thus incorporates the radial speed measurements as soon as they are available.
The increased robustness is demonstrated using robotic field experiment datasets collected using an all-terrain vehicle tailored for off-road navigation, driving at high speeds of up to \SI{\sim 12}{\meter\per\second}, in two different unstructured off-road environments. An instance of a high-speed driving in a desert environment is shown in \cref{fig:intro:figure}. In addition, the radial speed measurement factor is evaluated as a standalone odometry source and compared on a public dataset to state-of-the-art radar-inertial odometry approaches. Contributions of this work include:
\begin{itemize}
    \item A generalized method to integrate radar velocity measurements without depending on a least-squares solution, enabling the use of limited FoV single-beam radial velocity data at low-latency for reliable high-speed state estimation.
    \item High-speed robotic field tests conducted to demonstrate the increased robustness of the multi-modal state estimator when incorporating the proposed radar factor.
    \item Evaluation of proposed factor for standalone radial-inertial odometry estimation on public datasets and comparison with state-of-the-art methods.
\end{itemize}

The remainder of the manuscript is structured as follows. \Cref{sec:related_work} covers related work to multi-modal state estimation. \Cref{sec:method} describes the radial velocity factor and factor-graph based state estimator. \Cref{sec:evaluation} presents the experimental results, analysis, and comparisons. \cref{sec:conclusion} details the lessons learned and conclusions.

\section{Related Work}\label{sec:related_work}
Sensor fusion in state estimation remains as a core research topic for development of autonomous agents acting in complex environments under challenging conditions completing critical mission objectives. Many approaches take inspiration from classical GPS-aided inertial navigation, solving the problem of achieving both high accuracy and high rate state estimation, leveraging the complementary strengths of diverse sensors in order to mitigate their individual weaknesses. Although GPS has long been the go-to aiding sensor, many environments render its use not possible. Thus, a number of proposed approaches have explored other options for aiding sensors. With multi-sensor and LiDAR-centric \ac{slam} methods maturing, research has become more interested in investigating how aforementioned methods perform in challenging environments. In particularly how multiple, new sensors can be leveraged for better performance across a host of environmental conditions~\cite{cadena2016future,ebadi2024present}. Furthermore, high-speed autonomy in all-terrain conditions is not only interesting for automotive applications on Earth, but also required for pushing the limit of what is possible for extraterrestrial science mission objectives following~\cite{matthies2022longrange, rodriguez-martinez2019high-speed}.

For automotive research, LiDAR has become a key sensor, often paired with an IMU, for its vast utility, from state estimation to mapping, and high-accuracy and wide \ac{fov} data~\cite{levinson2011autonomous}.
While LiDAR odometry methods are numerous, maintaining a map and performing scan-to-map registration (commonly utilizing a variant of the \ac{icp} algorithm~\cite{besl1992icp,low2004point2plane}) remains a core step~\cite{zhang2014rss}.
Although, for all its strengths, LiDAR-based perception can still be challenged in environments containing obscurants or environments which are not well handled by odometry methods that rely on geometric alignment of point clouds. In~\cite{tuna2024xicp}, the authors propose methods for detecting degenerate axes resulting from a lack of diverse environmental geometry and integrate this in their point cloud registration. \ac{fmcw} radars are another typical automotive perception sensor, which have been gaining attention in the robotics community owing to their unique properties (e.g. directly measuring radial speed), with application possibilities in environments where more traditional perception (e.g. vision or LiDAR) struggle~\cite{harlow2024new}. Extreme environments aside, challenging environmental conditions can also be found even in typical automotive operation scenarios. To that end, the authors of~\cite{burnett2022weather} investigate the effects of weather on LiDAR perception and demonstrate the potential for increased performance with radar-to-radar and radar-to-LiDAR scan matching methods.

In terms of radar-inertial odometry many different methods have been proposed. The authors in \cite{kellner2013instant} first demonstrate the potential for ego-motion estimation using \ac{fmcw} radar sensors by robust least squares estimation. In~\cite{doer2022xrio}, the authors estimate linear velocity and yaw angle from \ac{fmcw} radar point clouds, and then fuse these estimates with IMU measurements in an EKF for aided inertial navigation. Opting for a more tightly-coupled approach, the authors in \cite{michalczyk2023tight} fuse the individual radial speed measurements along with point associations for aided inertial navigation. As opposed to the classical EKF-based approaches, the authors in \cite{kramer2021micro} construct a sliding-window smoother to solve the estimation optimization problem, also demonstrating performance in dense fog. Importantly, not all radar sensors are alike and spinning radars --which typically do not measure doppler-- are more popular in automotive applications; the authors of~\cite{adolfsson2023radar} propose a method for filtering and thereafter registration for high-accuracy scan-matching radar odometry.

Given their prevalence in automotive applications and the potential for increased robustness and accuracy derived from their complementary strengths, LiDAR-radar fusion methods have also been of research interest. The authors in~\cite{park2019radar} fuse LiDAR and spinning radar measurements by performing registration between radar point cloud measurements on a prior LiDAR map. Radar-to-LiDAR registration was also explored in~\cite{yin2022Rall}, where the authors developed an end-to-end machine learning method for performing localization through registration of a spinning radar point cloud on a LiDAR map. Our previous work~\cite{nissov2024roamer} investigated radar-LiDAR fusion for increased robustness in off-road environments utilizing a sliding-window smoother, with the disadvantage of only adding velocity information in the body-frame forward direction. In~\cite{nissov2024degradation}, the authors proposed a method for integrating LiDAR features in a factor graph smoother with radar least squares linear velocity estimates (both with the same measurement frequency) for improving the methods performance in environments with geometric self-similarity or dense fog. This method significantly changes the structure of the LiDAR factor, breaking the pose unary factor apart into individual feature factors. This change comes at a cost to overall compute, while maintaining the necessity that a single radar measurement has sufficient information to solve for the 3D linear velocity, which is not necessarily true for all radar sensors (e.g., the one utilized in this manuscript).

\section{Proposed Approach}\label{sec:method}
\subsection{Notation and Coordinate Frames}
This work utilizes multiple coordinate frames and the transformations between them, where the body-fixed ones are assumed to be known. The frames include the world frame denoted $\mathtt{W}$ and the body-fixed frames for the IMU, LiDAR, \ac{fmcw} radar, and GPS sensors denoted by $\mathtt{I}$, $\mathtt{L}$, $\mathtt{R}$, and $\mathtt{G}$.

Furthermore, let the position of frame $\mathtt{R}$ with respect to $\mathtt{I}$ expressed with respect to $\mathtt{W}$ be ${}_\mathtt{W}\bm{p}_\mathtt{IR}\in\mathbb{R}^3$ and let the rotation from $\mathtt{R}$ to $\mathtt{I}$ be ${}_\mathtt{I}\mathbf{R}_\mathtt{R}\in\textit{SO}(3)$. Furthermore, a translation and rotation combine to make the homogeneous transformation from $\mathtt{I}$ to $\mathtt{W}$: ${}_\mathtt{W}\mathbf{T}_\mathtt{I}\in\textit{SE}(3)$. Each aforementioned Lie group, $\textit{SO}(3)$ and $\textit{SE}(3)$, also has a Lie algebra which can be related to the element on the group by the exponential mapping
\begin{equation}
    \mathbf{R} = \exp{\bm{\tau}},\quad \mathbf{R}\in\textit{SO}(3),\ \bm{\tau}\in\mathfrak{so}(3)\hlf{,}
\end{equation}
using $\textit{SO}(3)$ \hlf{and its' algebra $\mathfrak{so}(3)$} as an example. Note, an element in the algebra $\bm{\tau}$ can be related to a Cartesian space through the \emph{wedge} and \emph{vee} mappings such that
\begin{equation}
\begin{aligned}
    (\cdot)^\wedge : \mathbb{R}^a \rightarrow \mathfrak{a}
    \hlf{,\quad 
    (\cdot)^\vee : \mathfrak{a} \rightarrow \mathbb{R}^a,}
\end{aligned}
\end{equation}
for an arbitrary Lie algebra $\mathfrak{a}$ with $a$ degrees of freedom. As an example, for $\textit{SO}(3)$ the wedge operator creates a skew symmetric matrix from a vector in $\mathbb{R}^3$. The interested reader can see~\cite{solà2021micro,barfoot2017state} for more in-depth coverage of Lie theory.

\subsection{\acl{fmcw} Radar} \label{sec:method:radar}
As an introduction to \ac{fmcw} radars, we provide a brief overview of the sensing principles relevant for the proposed method and refer the reader to~\cite{richards2013Fundamentals} for a comprehensive overview.

\ac{fmcw} radars function by transmitting a \emph{chirp}, which is a frequency modulated signal (typically modulated linearly with time), and processing the returns to reveal relative position and radial speed of a reflecting object. 

The time difference 
between when the chirp began transmitting and a return is received thus relates to the distance to the reflecting object.
This time difference manifests as a constant frequency in the mixed signal (made by combining a transmit-return pair). Thus, one can apply a \ac{fft}
to separate frequencies resulting from objects at different ranges from the sensor and thereby separate the range measurements, this operation is referred to as the range-\ac{fft}. By transmitting and receiving multiple chirps back-to-back, the doppler can be measured by taking another \ac{fft}, referred to as the doppler-\ac{fft}, and inspecting the phase difference 
of the same range detection peak represented in multiple chirps. This phase difference relates to the doppler measurement $v_r$ (also called radial speed) of a given object
as a function of the transmit time of a single chirp and the wavelength. Then, by transmitting chirps across an array of antennas, with known physical displacements, the azimuth and elevation angles for a given return can also be resolved.
Along a given bearing vector (calculated by normalizing Cartesian coordinates or from azimuth and elevation angles), the result of the range- and doppler-\acp{fft} can be restructured into a 2D array known as a \ac{rdmap}. The \ac{rdmap} is an image-like data structure filled with signal intensities, where rows correspond to range values and columns correspond to doppler values.

\begin{figure}[h!]
    \centering
    \includegraphics[width=0.9\linewidth]{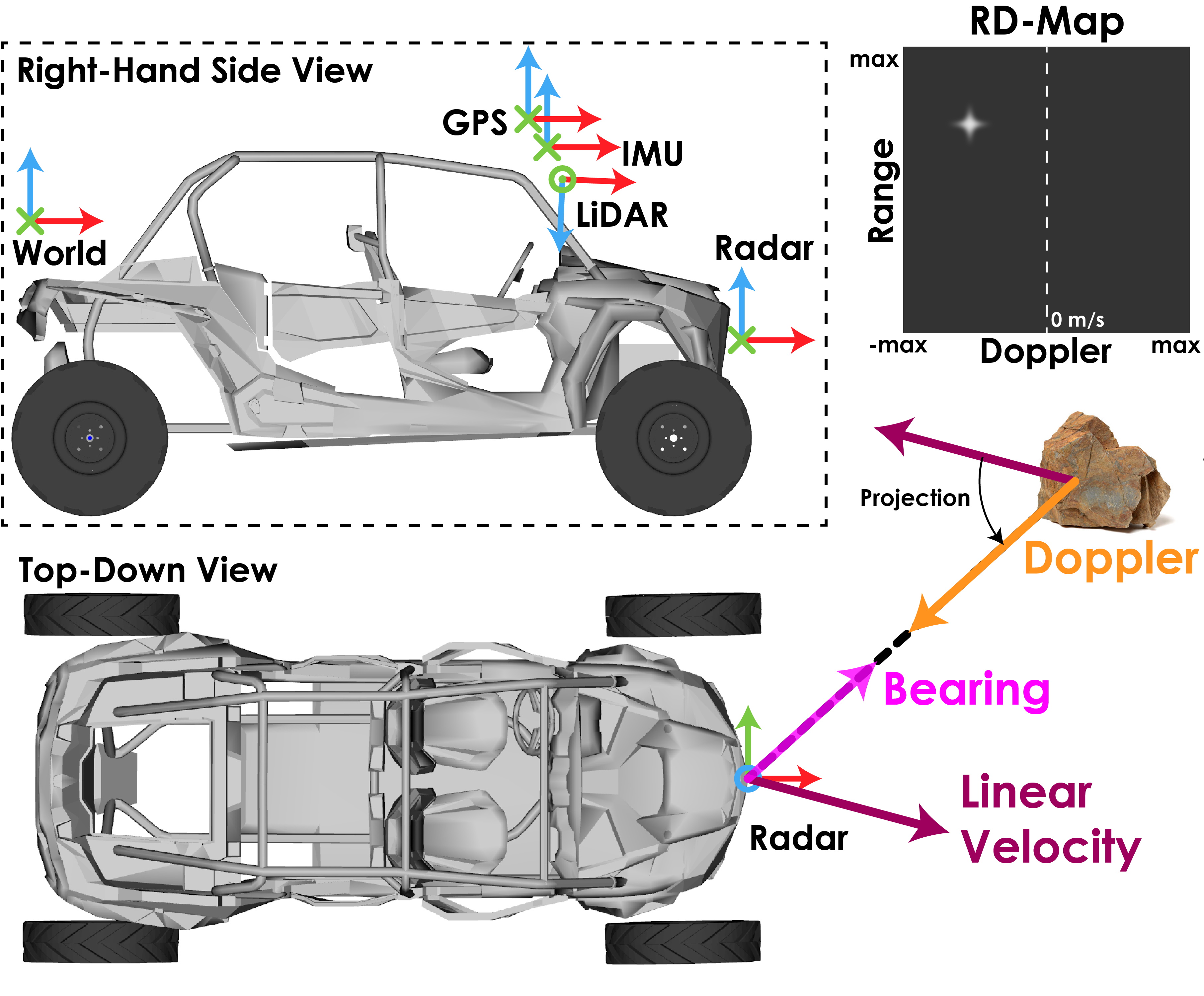}
    \caption{Visualization of the different sensor frames on the vehicle as well as an example of the radar return from a static object and the resulting \ac{rdmap}.}
    \label{fig:method:radial_speed}
\vspace{-1.5em}
\end{figure}

For estimation purposes, while the point cloud resulting from an \ac{fmcw} radar can be sparse and noisy~\cite{harlow2024new} and therefore difficult to use for registration, the doppler measurements relate trivially to the relative motion between the radar sensor and the environment. This is further compounded by sensors which return measurements covering a small FoV or with sparse returns (as in~\cite{nissov2024roamer}), meaning the radial speed is the more robust measurement as it can be fused instantaneously without consideration of previous measurements. Therefore, assuming a static environment, the radial speed measurement can be expressed as a function of the vehicle ego-motion, and used in aided inertial navigation applications, such that
\begin{equation}\label{eq:method:radial_speed}
    v_r = -{}_\mathtt{R}\bm{\mu}^\top {}_\mathtt{R}\bm{v}_\mathtt{WR}\hlf{,}
\end{equation}
for the radial speed along a unit-length bearing vector ${}_\mathtt{R}\bm{\mu}$, pointing towards the target's 3D location in $\mathtt{R}$, induced by the radar sensor's linear velocity ${}_\mathtt{R}\bm{v}_\mathtt{WR}$. A visual depiction of an \ac{rdmap} along with the relevant coordinate frames can be seen in \cref{fig:method:radial_speed}.

Although different radars can represent measurements in different forms (e.g. as point clouds, \acp{rdmap}, etc.) the aforementioned physical relationships hold. As such, the formulation for including the measurement in a factor graph, which is described in the following section, is applicable to virtually any \ac{fmcw} radar. The only requirement being that the radar provides doppler and bearing information. The latter can come either in the form of the unit bearing vector itself or azimuth and elevation angles. 
Note that since the factor is a function of the doppler and bearing, the range measurement itself is not necessary.

\vspace{-0.5em}
\subsection{State Estimation}
For the state estimation problem, let ${}_\mathtt{W}\mathbf{T}_\mathtt{I} \in \textit{SE}(3)$ be the homogeneous transformation from $\mathtt{I}$ to $\mathtt{W}$ expressed in $\mathtt{W}$, let ${}_\mathtt{W}\bm{v}_\mathtt{WI}$ be the linear velocity of $\mathtt{I}$ with respect to $\mathtt{W}$ expressed in $\mathtt{W}$, and let ${}_\mathtt{I}\bm{b}\in\mathbb{R}^6$ be the stacked accelerometer ($\bm{b}_{a}$) and gyroscope biases ($\bm{b}_{g}$) such that the state space is:
\begin{equation}
    \mathbf{x} = \begin{pmatrix}
        {}_{\mathtt{W}}\mathbf{T}_{\mathtt{I}} &{}_{\mathtt{W}}\bm{v}_{\mathtt{WI}} &{}_{\mathtt{I}}\bm{b}
    \end{pmatrix}\hlf{,}
\end{equation}
with the following dynamics derived for a locally tangent navigation frame
\begin{equation}
    \begin{aligned}
        {}_\mathtt{W}\dot{\mathbf{R}}_\mathtt{I} &= {}_\mathtt{W}\mathbf{R}_\mathtt{I} \left( {}_\mathtt{I}\tilde{\bm{\omega}}_{\mathtt{WI}} - \bm{b}_{g} \right)^\wedge\\
        {}_\mathtt{W}\dot{\bm{p}}_\mathtt{WI} &= {}_\mathtt{W}\bm{v}_{\mathtt{WI}}\\
        {}_\mathtt{W}\dot{\bm{v}}_{\mathtt{WI}} &= {}_\mathtt{W}\mathbf{R}_\mathtt{I} \left( {}_\mathtt{I}\tilde{\bm{a}}_\mathtt{WI} - \bm{b}_{a} \right) + {}_{\mathtt{W}} \bm{g}\\
        \dot{\bm{b}}_g &= \bm\nu_g\\
        \dot{\bm{b}}_a   &= \bm\nu_a\hlf{,}
    \end{aligned}
    \vspace{-0.5em}
\end{equation}
where ${}_\mathtt{I}\tilde{\bm{\omega}}_{\mathtt{WI}}$ and ${}_\mathtt{I}\tilde{\bm{a}}_\mathtt{WI}$ are the angular velocity and linear acceleration measured by the IMU, ${}_{\mathtt{W}}\bm{g}$ is the local gravity vector, and the random walk bias dynamics driven by zero-mean Gaussian noise sources $\bm\nu_g$ and $\bm\nu_a$. Note, the state estimation is grounded in the $\mathtt{I}$-frame, as such measurements with respect to other sensor frames must be related to physical quantities expressed with respect to $\mathtt{I}$.

\begin{figure}[h]
    \centering
    \includegraphics[width=0.9\linewidth]{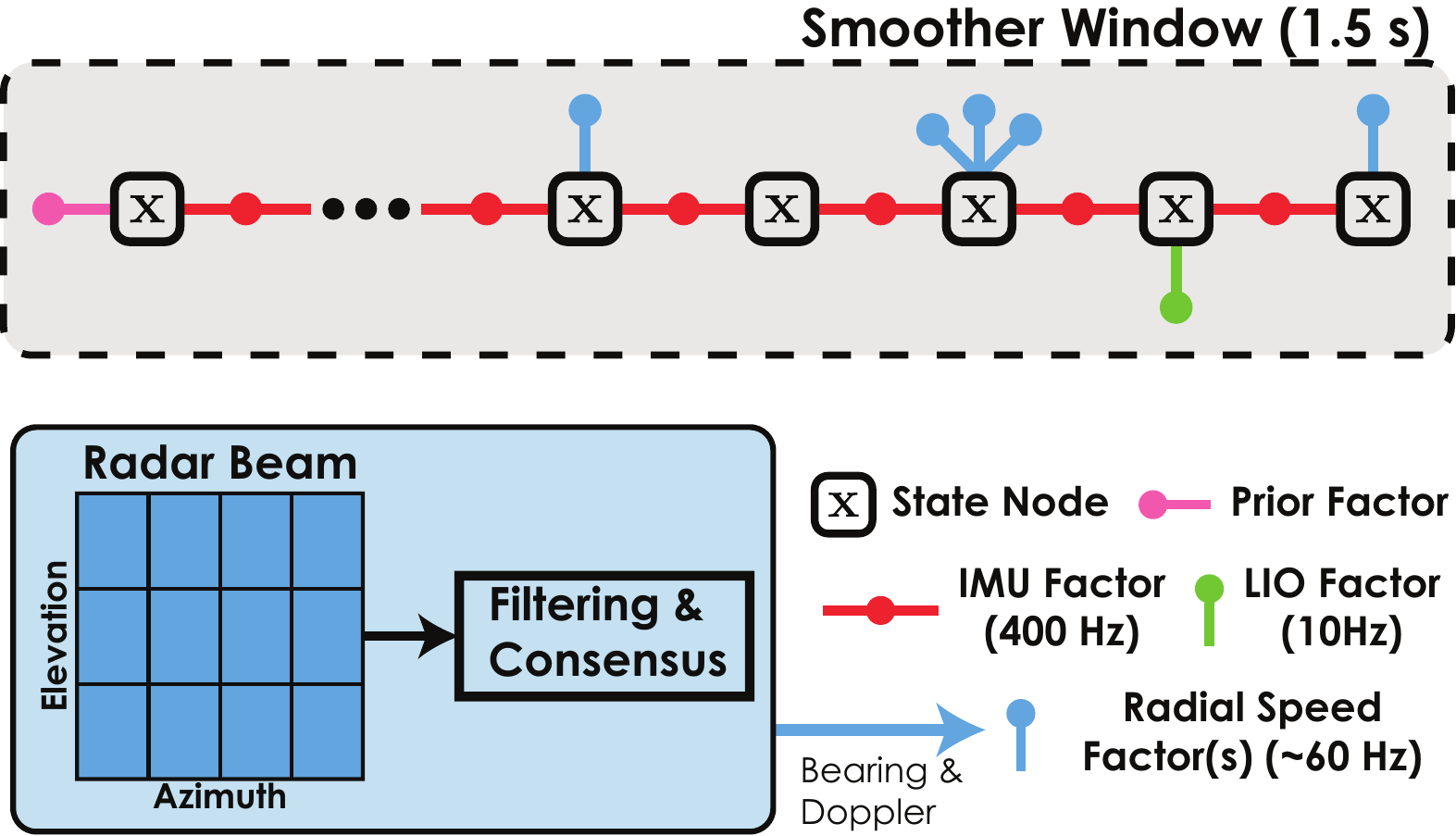}
    \caption{Illustration visualizing the connections in the factor graph, including factors from the IMU, LiDAR odometry, and radar connected to pose, linear velocity, and IMU bias states along with a marginalization prior for the sliding-window of the smoother.}
    \label{fig:method:graph_architecture}
    \vspace{-1em}
\end{figure}

Given the aforementioned state space, estimates can be constructed by correlating dynamics with sensors measurements. In the proposed method this is handled by a factor graph-based sliding-window smoother for the aided inertial navigation problem, where the aiding sensors are LiDAR (providing pose information) and \ac{fmcw} radar (providing linear velocity information). In terms of assembling the optimization problem (in the style of \cite{forster_-manifold_2017}), 
let $\mathcal{I}_i$ denote the IMU measurement sampled at time $i$, let $\mathcal{L}_j$ denote the LiDAR odometry (LO) measurement from time $j$, and let $\mathcal{R}_k$ denote the radar measurement from time $k$.
Furthermore, let the set of measurements times from the IMU, LiDAR, and radar collected from 0 up to time $m$ be ${}_{\mathtt{I}}{\mathcal{M}}_{0:m}$, ${}_{\mathtt{L}}{\mathcal{M}}_{0:m}$, and ${}_{\mathtt{R}}{\mathcal{M}}_{0:m}$, respectively.
The smoother thus calculates the optimal estimate for the set of states $\mathcal{X}_{m-l:m}$ (from $m$ to $m-l$ given by the smoother lag $l$), assuming zero-mean Gaussian noise models, by solving the following minimization problem
\begin{multline}
     {\mathcal{X}}_{m-l:m}^* = \argmin_{ {\mathcal{X}}_{m-l:m}} \Big( \lVert \bm{e}_{0} \rVert_{\bm\Sigma_0}^2 + \Sigma_{i\in{}_{\mathtt{I}}{\mathcal{M}}_{m-l:m}} \lVert \bm{e}_{{\mathcal{I}}_{i}} \rVert_{\bm\Sigma_{\mathcal{I}}}^2\\ + \Sigma_{j\in{}_{\mathtt{L}}{\mathcal{M}}_{m-l:m}}\lVert \bm{e}_{{\mathcal{L}}_j} \rVert_{\bm\Sigma_{\mathcal{L}}}^2 + \Sigma_{k\in{}_{\mathtt{R}}{\mathcal{M}}_{m-l:m}}\lVert \bm{e}_{{\mathcal{R}}_k} \rVert_{\bm\Sigma_{\mathcal{R}}}^2 \Big)\hlf{,}
\end{multline}
where $\bm{e}$ and $\bm{\Sigma}$ denote the residual and covariance matrix, with a subscript of $0$ corresponding to the marginalization prior and $\mathcal{I}$, $\mathcal{L}$, or $\mathcal{R}$ for values derived from sensor measurements. Note, the smoother window optimization and the marginalization prior is calculated according to~\cite{kaess2012smoothing,indelman2012smoothing}.
In the factor graph, see \cref{fig:method:graph_architecture} for the architecture associated with the proposed method, state estimates are denoted by nodes connected to each other by factors. These factors encode the information provided by the measurements as a function of the state space through the residual/covariance matrix pair.
The factors used by the proposed method are described in the following sections.

\subsubsection{Radar Factor}
Radar measurements are processed according to \cref{sec:method:radar} and the radial speed measurement from a given radar beam is paired with the corresponding beam-center bearing vector when calculating the relationship between the aforementioned measurement and the state space. The relationship between sensor linear velocity and radial speed along a bearing vector is given by \cref{eq:method:radial_speed}, although this is not yet in terms of the state space. Radar linear velocity is related to IMU linear velocity by taking into consideration the effect of a rotating reference frame (as in \cite{doer2020ekfrio}), such that
\begin{equation}
    {}_\mathtt{R}\bm{v}_\mathtt{WR} = {}_\mathtt{R}\mathbf{R}_\mathtt{I} \left( {}_\mathtt{I}\bm{v}_\mathtt{WI} + {}_\mathtt{I}\bm{\omega}_\mathtt{WI} \times {}_\mathtt{I}\bm{p}_\mathtt{IR} \right)\hlf{,}
\end{equation}
using the IMU-radar extrinsic transformation $\{{}_\mathtt{R}\mathbf{R}_\mathtt{I},\ {}_\mathtt{I}\bm{p}_\mathtt{IR}\}\in\textit{SE}(3)$. This can be expanded in terms of variables in the state space and IMU measurements as
\begin{equation}
    {}_\mathtt{R}\bm{v}_\mathtt{WR} = {}_\mathtt{R}\mathbf{R}_\mathtt{I} \left( {}_\mathtt{W}\mathbf{R}_\mathtt{I}^\top {}_\mathtt{W}\bm{v}_\mathtt{WI} + \left( {}_\mathtt{I}\tilde{\bm{\omega}}_\mathtt{WI} - \bm{b}_g \right) \times {}_\mathtt{I}\bm{p}_\mathtt{IR} \right)\hlf{,}
\end{equation}
such that the residual error of a given radar measurement (composed of doppler and bearing vector) can be written as
\begin{equation}
    \bm{e}_\mathcal{R} = -{}_\mathtt{R}\bm{\mu}^\top {}_\mathtt{R}\mathbf{R}_\mathtt{I} \left( {}_\mathtt{W}\mathbf{R}_\mathtt{I}^\top {}_\mathtt{W}\bm{v}_\mathtt{WI} \hlf{+} \left( {}_\mathtt{I}\tilde{\bm{\omega}}_\mathtt{WI} - \bm{b}_g \right) \times {}_\mathtt{I}\bm{p}_\mathtt{IR} \right) - \tilde{v}_r\hlf{,}
\end{equation}
which defines the factor implementation along with the corresponding Jacobians with respect to the state space members. The non-zero Jacobians for the radar radial speed factor are
\begin{equation}
\begin{aligned}
    \frac{\partial e_\mathcal{R}}{\partial {}_\mathtt{W}\mathbf{R}_\mathtt{I}} &= -{}_\mathtt{R}\bm{\mu}^\top {}_\mathtt{R}\mathbf{R}_\mathtt{I} \left( {}_\mathtt{I}\mathbf{R}_\mathtt{W} {}_\mathtt{W}\bm{v}_\mathtt{WI} \right)^\wedge\\
    \frac{\partial e_\mathcal{R}}{\partial {}_\mathtt{W} \bm{v}_\mathtt{WI}} &= -{}_\mathtt{R}\bm{\mu}^\top {}_\mathtt{R}\mathbf{R}_\mathtt{I} {}_\mathtt{I}\mathbf{R}_\mathtt{W}\\
    \frac{\partial e_\mathcal{R}}{\partial \bm{b}_g} &= -{}_\mathtt{R}\bm{\mu}^\top {}_\mathtt{R}\mathbf{R}_\mathtt{I} \left( {}_\mathtt{I}\bm{p}_\mathtt{IR} \right)^\wedge
    .
\end{aligned}
\end{equation}
Outlier measurements are handled by applying a Huber loss to the radar factor in the factor graph.

\subsubsection{IMU Factor}
The IMU factor used by the proposed method is the pre-integration factor proposed in~\cite{forster_-manifold_2017}, which includes residual functions taken with respect to changes in the attitude $\bm{e}_{\Delta{}_{\mathtt{W}}{\mathbf{R}}_{\mathtt{I}}}$, position $\bm{e}_{\Delta{}_{\mathtt{W}}\bm{p}_{\mathtt{WI}}}$, and velocity $\bm{e}_{\Delta{}_{\mathtt{W}}\bm{v}_{\mathtt{WI}}}$ such that the total IMU residual is
\begin{equation}
    \bm{e}_{\mathcal{I}} = 
    \begin{bmatrix}
        \bm{e}_{\Delta{}_\mathtt{W}\mathbf{R}_\mathtt{I}}^\top &\bm{e}_{\Delta{}_\mathtt{W}\bm{p}_\mathtt{WI}}^\top &\bm{e}_{\Delta{}_\mathtt{W}\bm{v}_\mathtt{WI}}^\top
    \end{bmatrix}^\top
    .
\end{equation}
Note, differently from~\cite{forster_-manifold_2017}, this factor graph construction takes after~\cite{nubert2022graph}, where nodes are added to the graph at every IMU message. As a result this factor connects nodes with a single IMU measurement, as opposed to lumping together multiple into a single factor. This allows for easy integration of delayed measurements by association with the closest node in the graph by timestamp. As the nodes are distributed according to the IMU sampling rate, the error committed by this association is minimized.

\subsubsection{LiDAR Odometry Measurements}
In the absence of absolute position information, e.g. GPS, LiDAR Odometry (LO) estimates are added to the factor graph as pseudo global pose updates in the form of unary factors.
The LO method takes as input IMU measurements and multiple LiDAR point clouds, three in our experiments (Section~\ref{sec:experiment:jpl}). 
The input point clouds are undistorted at a common timestamp using IMU measurements and expressed in a common LiDAR frame ($\mathtt{L}$) using extrinsic calibrations. The merged point cloud is used to perform a scan--to-submap registration to estimate the robot pose ${}_\mathtt{W}\hat{\mathbf{T}}_\mathtt{I}$, as detailed in~\cite{fakoorian2022rose}. The residual error for the pose--unary factor is given as:
\begin{equation}
    \bm{e}_\mathcal{L} = \log \left( \left( {}_\mathtt{W}\hat{\mathbf{T}}_\mathtt{I} \right)^{-1} {}_\mathtt{W}\mathbf{T}_\mathtt{I} \right)^{\hlf{\vee}} \hlf{,}
\end{equation}
where $\log$ is the logarithmic mapping from $\textit{SE}(3)$ to $\mathfrak{se}(3)$.

\section{Experimental Evaluation}\label{sec:evaluation}
The performance of proposed method for high-speed off-road robotic navigation was evaluated by conducting real-world field experiments. Furthermore, a comparison with the current state-of-the-art radar-inertial state estimation methods is presented using the public Forest and Mine dataset~\cite{kubelka2024need}.

\subsection{High-speed Off-road Robot Experiments}\label{sec:experiment:jpl}
The proposed method was experimentally tested at the NASA \ac{jpl} using a modified Polaris RZR all-terrain vehicle, shown in \cref{fig:intro:figure}. During experiments, sensor data from a Xsens MTi-630 IMU (\SI{400}{\hertz}), three Velodyne VLP-32 LiDARs (\SI{10}{\hertz}), an Echodyne EchoDrive \ac{fmcw} radar, and a Swift Navigation Duro GPS was collected. All sensors are synchronized to the GPS time through PTP.

The EchoDrive radar measures a grid of \acp{rdmap} along a narrow bearing window. This window is referred to as a \emph{beam}, composed of $3\times4$ \acp{rdmap} each spanning \SI[round-precision=0]{1}{\degree} of azimuth and \SI[round-precision=1]{2.5}{\degree} of elevation. A single doppler/bearing pair are extracted from each beam measurement, following the square-law filtering in~\cite{nissov2024roamer}, and added to the graph with a single radial factor.
The radar 
parameters used in the experiments are presented in \cref{tab:evaluation:radar_chirp}
\begin{table}[ht]
    \centering
    \caption{Parameters for the {S21a} EchoDrive radar chirp.}
    \label{tab:evaluation:radar_chirp}
    \sisetup{
        table-format=2.4,
        table-alignment-mode=format,
        round-precision=3
    }
    \vspace{-1ex}
    \begin{tabular}{ll}
        \toprule
        Parameter   &Value\\
        \midrule
        Max Range   &\SI{100}{\meter}\\
        Range Resolution   &\SI{0.49}{\meter}\\
        Max Doppler   &\SI{43.178}{\meter\per\second}\\
        Doppler Resolution   &\SI{0.169}{\meter\per\second}\\
        Beam Sampling Time   &\SI{15.8}{\milli\second}\\
        \bottomrule
    \end{tabular}
    \vspace{-2ex}
\end{table}
\begin{table*}[htb]
    \centering
    \caption{Root Mean Square Errors (RMSE) and standard deviations of pose and velocity estimates for the JPL East Lot off-road track and Helendale desert experiments.}
    \label{tab:evaluation:jpl:results}
    \sisetup{
        table-format=2.4,
        table-alignment-mode=format,
        round-precision=3
    }
    \hlf{
    \begin{tabular}{lllcccccc}
         \toprule
         &\multicolumn{2}{l}{\multirow{2}{*}{Method}}   &\multicolumn{2}{c}{APE RMSE}    &\multicolumn{2}{c}{RPE RMSE ($\Delta$:\SI{10}{\meter})}   &\multicolumn{2}{c}{Velocity [\si{\meter\per\second}]}\\
         &\multicolumn{2}{l}{}  &{Translation [\si{\meter}]}    &{Rotation [\si{\degree}]}   &{Translation [\si{\meter}]}    &{Rotation [\si{\degree}]}   &{Forward}    &{Lateral}\\
         \midrule
        \multirow{12}{*}{East Lot} &\multicolumn{2}{l}{LI}    &$2.39 \pm 0.90$	&$1.33 \pm 0.57$	&$0.20 \pm 0.12$	&$0.35 \pm 0.30$	&$0.06 \pm 0.06$	&$0.13 \pm 0.13$\\
        &\multicolumn{2}{l}{LRI}  &$2.26 \pm 0.83$	&$1.96 \pm 0.81$	&$0.20 \pm 0.10$	&$0.78 \pm 0.35$	&$0.07 \pm 0.06$	&$0.15 \pm 0.13$\\ \cmidrule{2-9}
        &\multirow{5}{*}{Noisy-LI} &$\sigma_{xy}=1$  &$2.50 \pm 0.99$	&$2.50 \pm 1.07$	&$0.79 \pm 0.36$	&$2.51 \pm 1.23$	&$0.42 \pm 0.42$	&$0.46 \pm 0.46$\\
        &  &$\sigma_{xy}=2$  &$3.04 \pm 1.19$	&$4.05 \pm 1.88$	&$1.56 \pm 0.72$	&$4.19 \pm 1.94$	&$0.77 \pm 0.77$	&$0.82 \pm 0.82$\\
        &  &$\sigma_{xy}=4$  &$4.33 \pm 1.99$	&$7.25 \pm 3.50$	&$2.60 \pm 1.22$	&$5.86 \pm 2.61$	&$1.37 \pm 1.37$	&$1.50 \pm 1.50$\\ \cmidrule{3-9}
        &  &$\sigma_{xy}=\sqrt{\lVert {}_\mathtt{I}\bm{v}_\mathtt{WI} \rVert_2}$  &$3.38 \pm 1.49$	&$5.45 \pm 2.72$	&$2.02 \pm 0.90$	&$4.97 \pm 2.19$	&$1.02 \pm 1.02$	&$1.12 \pm 1.12$\\ \cmidrule{2-9}
        &\multirow{5}{*}{Noisy-LRI}     &$\sigma_{xy}=1$  &$2.31 \pm 0.85$	&$2.38 \pm 1.19$	&$0.48 \pm 0.25$	&$1.85 \pm 0.92$	&$0.07 \pm 0.06$	&$0.15 \pm 0.13$\\
        &  &$\sigma_{xy}=2$  &$3.16 \pm 1.36$	&$3.35 \pm 1.92$	&$0.84 \pm 0.48$	&$3.04 \pm 1.72$	&$0.07 \pm 0.06$	&$0.15 \pm 0.14$\\
        &  &$\sigma_{xy}=4$  &$4.07 \pm 1.91$	&$4.51 \pm 2.65$	&$1.33 \pm 0.77$	&$4.44 \pm 2.54$	&$0.07 \pm 0.06$	&$0.15 \pm 0.14$\\ \cmidrule{3-9}
        &  &$\sigma_{xy}= \sqrt{\lVert {}_\mathtt{I}\bm{v}_\mathtt{WI} \rVert_2}$  &$3.13 \pm 1.31$	&$3.78 \pm 2.23$	&$1.09 \pm 0.64$	&$3.71 \pm 2.12$	&$0.07 \pm 0.06$	&$0.15 \pm 0.14$\\
        \midrule
        Helendale  &\multicolumn{2}{l}{LRI}  &$18.00 \pm 11.16$	&$2.33 \pm 0.75$	&$0.44 \pm 0.19$	&$0.90 \pm 0.46$	&$0.07 \pm 0.06$	&$0.22 \pm 0.15$\\
        \bottomrule
    \end{tabular}
    }
    \vspace{-0.5em}
\end{table*}
\begin{figure*}
     \centering
     \subfloat[North and east position estimates.]{\includegraphics[width=0.33\textwidth]{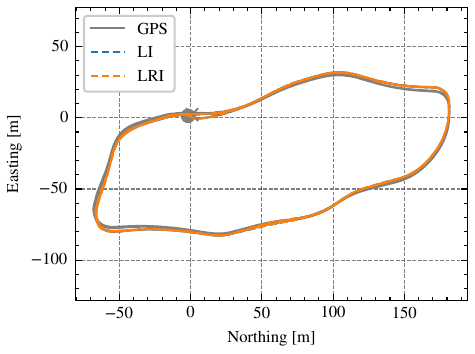}\label{fig:evaluation:east_lot5:xy}}\hfill
     \subfloat[Body-frame forward velocity error.]{\includegraphics[width=0.33\textwidth]{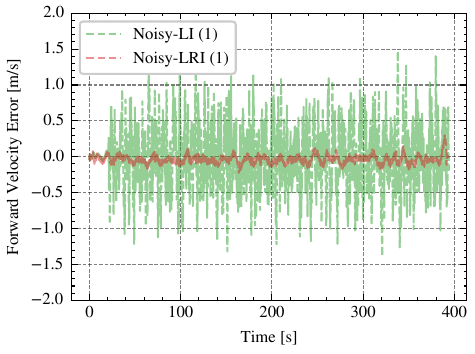}\label{fig:evaluation:east_lot5:vx_err}}
     \hfill
     \subfloat[Body-frame lateral velocity error.]{\includegraphics[width=0.33\textwidth]{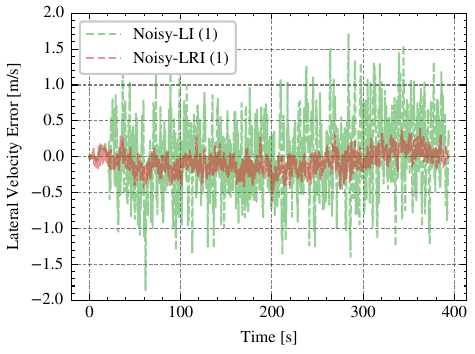}\label{fig:evaluation:east_lot5:vy_err}}
     \caption{Figures comparing the performance of different methods versus GPS in the \ac{jpl} East Lot experiment including nominal performance (a) as well as the added noise experiments (b,c) where the noise standard deviation is \SI{1}{\meter}.}
     \vspace{-1.5em}
\end{figure*}

For datasets collected with the \ac{jpl} vehicle (i.e., \Cref{sec:experiment:jpl:east_lot,sec:experiment:jpl:helendale}), the onboard GPS was fused with IMU to create a ground truth estimate for comparison purposes. Estimates for vehicle pose, linear velocity, and IMU bias were calculated using a factor graph, similar in architecture to the proposed method but with only IMU and GPS, using a Levenberg–Marquardt optimizer to solve for the state estimates of the full trajectory instead of a sliding-window optimization.

\subsubsection{Manual Off-road Driving Validation Experiment}\label{sec:experiment:jpl:east_lot}
To validate the performance of the proposed method, a dataset was collected using the \ac{jpl} vehicle driving 5 loops, in manual mode, on an off-road track near the \ac{jpl} campus' East Lot (the same track as in \cite{nissov2024roamer}). The total distance driven is approximately \SI{3.002133499199908}{\kilo\meter} long with the experiment lasting about \SI{6.479866667}{\minute}. The start and end positions of the vehicle at the end of completing 5 loops around the track are approximately the same. The vehicle drives through the course with an average speed of \SI{7.72169566}{\meter\per\second} and reaches a max speed of \SI{12.266}{\meter\per\second}. From this experiment, two investigations were conducted to analyze the performance of the proposed method with nominal and intentionally degraded LiDAR odometry.

\paragraph{Nominal Conditions} Nominal conditions for this experiment refer to the fact that the environment provided enough geometric features for the LiDAR odometry methods to perform accurately without experiencing any degradation in performance. As such, the pose estimates from the LO method provide a baseline to investigate if the addition of the radar information incurs any detrimental effects to the overall state estimation quality. 
For this experiment, qualitative and quantitative results are presented in \cref{fig:evaluation:east_lot5:xy} and top row of \cref{tab:evaluation:jpl:results}, respectively, and show that the system augmented with radar (denoted \emph{LRI}) demonstrates comparable performance to the proposed method without radar (denoted \emph{LI}) confirming that the proposed method does not effect the state estimation quality under nominal conditions in any adverse manner.

Note, in this experiment, optimization times were on average $14.612 \pm 0.097$~\textrm{ms} for LRI and $16.211\pm0.0254$~\textrm{ms} for LI, when run on a laptop with an Intel \nth{11} i7-11850H. Showing that on a CPU significantly less powerful than the off-road vehicles', the additional information from the radar serves to reduce time taken for single optimizations.

\paragraph{Degraded LiDAR Odometry} To demonstrate the robustness improvement due to the inclusion of radar velocity information on the state estimation process,
the LiDAR odometry factors are intentionally perturbed with noise before being added to the state estimator's factor graph. In the absence of radar factors, the addition of noise makes the state estimator output noisy, which further feeds back through the point cloud undistortion into the LiDAR odometry module. For all trials noise was added after an initial delay at the start of the run to allow for proper convergence of the IMU biases.
Furthermore, as radar velocity factors only provide linear velocity information along $x$ and $y$ axes, Gaussian noise was only added along these axes to LO position estimates, with $\sigma_{xy}$, to understand the direct benefit of the improved velocity estimation in isolation.
Thus for each LO pose estimate, 
the noisy estimate ${}_{\mathtt{W}}\tilde{\mathbf{T}}_{\mathtt{I}}$ is given by
\begin{equation}
\begin{aligned}
    {}_{\mathtt{W}}\tilde{\mathbf{T}}_{\mathtt{I}} 
    &= {}_{\mathtt{W}}\hat{\mathbf{T}}_{\mathtt{I}} \exp\left( \bm\epsilon^\wedge \right),\ \bm\epsilon\in\mathcal{N}\left(\bm{0}, \Sigma \right)\hlf{,}
\end{aligned}
\end{equation}
where $\Sigma$ is the noise covariance.

Note, the factor graph parameters were not changed to take this added noise into consideration, the desire is to emulate the effects of unexpected worsening of LO quality on the state estimation process. Different values of noise were evaluated and the results are shown in \cref{tab:evaluation:jpl:results}, the results for $\sigma_{xy}=1$ are visualized in \cref{fig:evaluation:east_lot5:vx_err,fig:evaluation:east_lot5:vy_err}, where the radar (being integrated as a velocity sensor) has a substantial impact on improving the velocity estimation error, such that RMSE remains nearly constant regardless of the noise added. Furthermore, it can be noted in \cref{tab:evaluation:jpl:results}, that this information was able to improve the pose estimate quality as well, except for in a single instance where the \ac{ape} is a standout. This resulted from an offset in the $z$ estimate error for LRI which was not as significantly different from LI in the other noise combinations.

\subsubsection{Off-road Driving in Challenging Desert Environment}\label{sec:experiment:jpl:helendale}
\begin{figure*}
     \centering
     \subfloat[North and east position estimates.]{\includegraphics[width=0.33\textwidth]{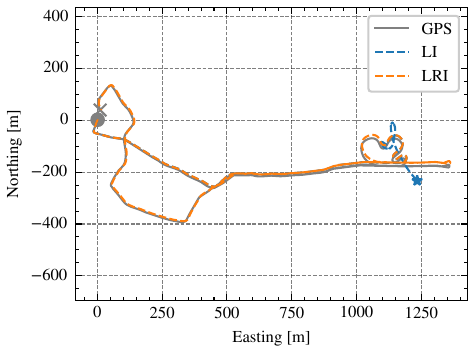}\label{fig:evaluation:helendale:xy}}\hfill
     \subfloat[Body-frame forward velocity estimates.]{\includegraphics[width=0.33\textwidth]{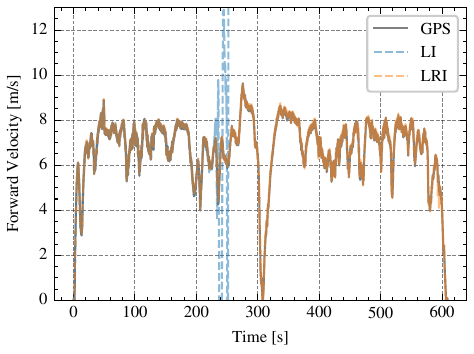}\label{fig:evaluation:helendale:vx}}\hfill
     \subfloat[Body-frame lateral velocity estimates.]{\includegraphics[width=0.33\textwidth]{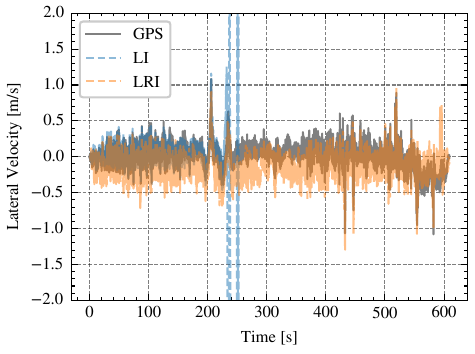}\label{fig:evaluation:helendale:vy}}
     \caption{Figures comparing the performance of different methods versus GPS in the \ac{jpl} Helendale experiment including position (a), forward velocity in body-frame (b), and lateral velocity in \hlf{body-frame estimates (c).}}
     \label{fig:evaluation:helendale}
     \vspace{-1.0em}
\end{figure*}
To test the applicability of the proposed method for real-world off-road driving scenarios, a test was conducted in a desert environment near Helendale, California, USA. For this test the \ac{jpl} vehicle was driven for a path length of approximately \SI[round-precision=0]{4.1008347915619315}{\kilo\meter} in a duration of approximately \SI[round-precision=0]{10.08035}{\minute}. During the experiment the average speed of vehicle was \SI[round-precision=1]{6.780245381}{\meter\per\second} with a maximum speed of \hlf{\SI[round-precision=1]{9.6}{\meter\per\second}}. The driven path goes through a relatively flat, desert environment containing sparse, low-density vegetation. As such, this environment has limited geometric features for LiDAR scan matching, aside from those found on the ground. In addition, the trajectory contains aggressive driving with maneuvers approaching \hlf{\SI{46}{\degree\per\second}} of yaw rate, as measured by the IMU.

This environment poses difficulties for a purely LiDAR-inertial method, as the flat regions with sparse vegetation do not provide sufficiently unique geometry. Furthermore, this challenge is compounded by the high-speed traversal with aggressive turns performed in said regions of the environment.
For these reasons, the LO performance degrades significantly and diverges during the most difficult passage of the course, as shown in \cref{fig:evaluation:helendale:xy}. For the same trajectory, addition of the radar velocity \hlf{information demonstrates} the increased robustness, as the additional information is able to stabilize the optimization and aid in \hlf{supporting} the LiDAR odometry during this particularly difficult region. \hlf{The result of which being} the proposed method keeping a reliable state estimate throughout the entire trajectory and tracking the GPS position of the vehicle\hlf{, with performance shown in \cref{tab:evaluation:jpl:results,fig:evaluation:helendale}}. Note, during the challenging portions, the estimation quality along $z$ did suffer, due to the fact that the radar does not provide significant information along this axis.
\vspace{-0.5em}

\subsection{Radar-Inertial Odometry Comparison with Public Dataset}
To demonstrate the suitability of the proposed radar velocity factor for radar-inertial odometry estimation as a standalone solution in challenging environments, comparisons with state-of-the-art radar-inertial methods were made on the publicly available \emph{Forest and Mine Datasets}~\cite{kubelka2024need}, which utilizes another automotive radar sensor (Sensrad Hugin A3).
The Forest dataset contains a vehicle driving through a forest environment on a path of \SI[round-precision=1]{1.8}{\kilo\meter} over a duration of \SI[round-precision=1]{8.3}{\minute}. The vehicle drives along the path at an average speed of \SI{\sim3}{\meter\per\second} and reaches a maximum speed of \SI{\sim6.5}{\meter\per\second}. The Mine dataset has a different vehicle driving through a mine for \SI{13}{\minute} on a path \SI[round-precision=1]{4.5}{\kilo\meter} in length. Here the vehicle reaches an average speed of \SI{\sim5}{\meter\per\second} with a maximum at \SI{\sim10}{\meter\per\second}.

A key difference between the Forest and Mine dataset and our experiments conducted with the \ac{jpl} vehicle, are the type of the radar sensors used. The radar used in~\cite{kubelka2024need} returns a processed point cloud, with radial speed measurements per point, reaching sizes of up to 10k points covering the full sensor FoV. As a result, the data must be processed differently before adding factors to the state estimator factor graph. Furthermore, this dataset features more busy environments and as such, the first step in the processing is to find the radar points originating from static objects in the environment. Mirroring the approach presented in~\cite{kubelka2024need}, assuming the consensus set is the set of static objects, we apply RANSAC for solving the linear velocity least squares problem relating the radial speed and bearing measurement with radar linear velocity:
\begin{equation}
    \begin{bmatrix}
        \tilde{v}_r^1\\
        \tilde{v}_r^2\\
        \vdots\\
        \tilde{v}_r^N
    \end{bmatrix}
    =
    \begin{bmatrix}
        \left( -{}_\mathtt{R}\tilde{\bm{\mu}}^1 \right)^\top\\ 
        \left( -{}_\mathtt{R}\tilde{\bm{\mu}}^2 \right)^\top\\
        \vdots\\
        \left( -{}_\mathtt{R}\tilde{\bm{\mu}}^N \right)^\top
    \end{bmatrix}
    {}_\mathtt{R}\bm{v}_\mathtt{WR}\hlf{,}
\end{equation}
where $\tilde{v}_r^n$ and ${}_\mathtt{R}\tilde{\bm{\mu}}^n$ are the doppler and bearing vector corresponding to the $n$th point. Afterwards, a subset of $N$ best points (according to error from estimated model) are selected from the inlier set.
Furthermore, points are iteratively selected to ensure new points maintain a minimum azimuth and elevation difference from the previously selected points, ensuring \hlf{a good distribution} over the sensor FoV.
\begin{table}
    \centering
\begin{threeparttable}
    \caption{Translation and rotation median RPE for different deltas in translation, comparing proposed with results from~\cite{kubelka2024need}.}
    \label{tab:evaluation:forest_and_mine}
    \sisetup{
        table-format=2.4,
        table-alignment-mode=format,
        round-precision=2
    }
    \begin{tabular}{llSSSS}
        \toprule
            &   &\multicolumn{2}{c}{Translation [\si{\percent}]}   &\multicolumn{2}{c}{Rotation [\si{\degree}]}\\
            &   &{$\Delta$:\SI{1}{\meter}}    &{$\Delta$:\SI{10}{\meter}}   &{$\Delta$:\SI{1}{\meter}}    &{$\Delta$:\SI{10}{\meter}}\\
        \midrule
        \multirow{3}{*}{\rotatebox[origin=c]{90}{Ours}}
            &Forest   &\num{11.539830197638775}   &\num{1.869877527662993}   &\num{0.271305854015442}   &\num{0.36446744046317725}\\
            &Mine   &\num{11.503457754416814}   &\num{2.0305090091523468}   &\num{0.2805072610020319}   &\num{0.37953909444527195}\\ \cmidrule{2-6}
            &Average    &\num{11.521643976}   &\num{1.950193268}   &\num{0.275906558}   &\num{0.372003267}\\
        \midrule
        \multirow{5}{*}{\rotatebox[origin=c]{90}{\makecell{Results\\ from~\cite{kubelka2024need}}}}
            &IMU+Doppler\tnote{a}~\cite{kubelka2024need}   &\num{9.99}   &\num{1.53}   &\num{0.19}   &\num{0.28}\\
            &EKF~\cite{doer2020ekfrio} &\num{34.54}   &\num{10.29}   &\num{0.29}   &\num{0.65}\\
            &ICP~\cite{kubelka2024need} &\num{32.68}   &\num{4.74}   &\num{0.83}   &\num{1.34}\\
            &APDGICP~\cite{zhang2023radarslam} &\num{33.83}   &\num{11.53}   &\num{0.76}   &\num{3.39}\\
            &NDT~\cite{zhang2023radarslam} &\num{27.70}   &\num{11.79}   &\num{0.56}   &\num{1.95}\\
        \bottomrule
    \end{tabular}
    \begin{tablenotes}
        \item[a] This method utilizes the magnetometer-aided attitude solution provided by the IMU to integrate radar velocity measurements. Therefore additional sensor information as compared to the other methods.
    \end{tablenotes}
\end{threeparttable}
\vspace{-2.5em}
\end{table}

Odometry performance is reported for the Forest and Mine datasets using median RPE as the metric, therefore we will compare the performance of the proposed method on these datasets in terms of median RPE evaluated of each of the two datasets and averaged together as well. Results for other methods~\cite{kubelka2024need,doer2020ekfrio,zhang2023radarslam}, performance metrics taken from~\cite{kubelka2024need}, along with results from the proposed method on the dataset can be seen in \cref{tab:evaluation:forest_and_mine}. Note, that this comparison is utilizing the proposed method's radial speed factor for radar-inertial odometry, which outperforms most methods used for comparison by the authors in~\cite{kubelka2024need}. In particular the comparison with the \emph{EKF}~\cite{doer2020ekfrio} solution is interesting, this is a reasonable comparison as both methods fuse velocity information from the radar sensors with accelerometer and gyroscope measurements from an IMU. The proposed method is only beat by \emph{IMU+Doppler} from~\cite{kubelka2024need}, which is a method that utilizes the magnetometer as well to limit orientation drift and integrates radar velocity measurements according to the IMU's orientation estimation algorithm. As such, this method considers additional information which \hlf{it} does not have access to, and even then the proposed method has comparable RPE. In conclusion, the proposed method is able to create smooth, locally accurate estimates, competing with other radar-inertial methods. \hlf{However, long-term drift, resulting} from error in yaw or $z$ as shown in~\cite{kubelka2024need,nissov2024degradation}, is still inevitable with only body-frame velocity and IMU information.

\section{Conclusions and Lessons Learned}\label{sec:conclusion}
These investigations have imparted several takeaways on the authors. The first of which is that sub-optimal or even degenerate conditions can appear suddenly, as was the case in the Helendale dataset. Estimation algorithms not equipped to deal with such unexpected scenarios can and will suffer in performance. Furthermore, utilizing complementary multi-modal measurement sources seem to be a valid methodology for creating a robust state estimation architecture. 
However, such fusion methodologies can, at times, return slightly surprising results; e.g. in the East Lot datasets despite the radar sensor presenting virtually no information regarding vertical velocity still manages to affect the overall result. Unfortunately, this also means that inaccuracy along the vertical channel has little potential to be improved in these experiments, as the beam distribution was not designed with a large vertical \ac{fov} in mind and as automotive radars in general tend to have a more limited vertical \ac{fov}. This matches the behavior observed in \cite{nissov2024roamer,nissov2024degradation}, as what seems to be a limitation of velocity-inertial fusion with \ac{fov} constrained radar sensing.

With that in mind, this work proposed fusion of radar velocity measurements in a factor graph optimization, in addition to LiDAR odometry and IMU measurements, to improve the state estimator robustness for robotic operation in complex environments under challenging conditions. The method was evaluated on off-road high-speed driving datasets under the influence of both real-world and artificial sensor degradation. Furthermore, the radial speed factor was evaluated as a standalone radar-inertial odometry method on a public dataset containing complex underground and forest environments and with a different \ac{fmcw} radar sensor. Improved robustness was demonstrated for off-road field experiments, and state-of-the-art comparative performance was shown on the public datasets.
\vspace{-1.0em}

\addtolength{\textheight}{-12cm}
\bibliographystyle{IEEEtran}
\bibliography{bib/general, bib/comparison}

\end{document}